\pdfoutput=1

\documentclass[11pt]{article}

\usepackage[final]{acl}

\usepackage{times}
\usepackage{latexsym}

\usepackage[T1]{fontenc}

\usepackage[utf8]{inputenc}

\usepackage{microtype}

\usepackage{inconsolata}

\usepackage{lipsum}

\usepackage{graphicx}
\usepackage{adjustbox}
\usepackage{amssymb}
\usepackage{amsmath}
\usepackage{multirow}
\usepackage{xspace}
\usepackage{booktabs}
\usepackage{hyperref}
\usepackage{subfig}
\usepackage{tikz}
\usetikzlibrary{shapes, arrows.meta, positioning}
\usepackage{array}
\newcolumntype{L}{>{\raggedright\arraybackslash}m{3cm}}
\title{\includegraphics[width=0.05\textwidth]{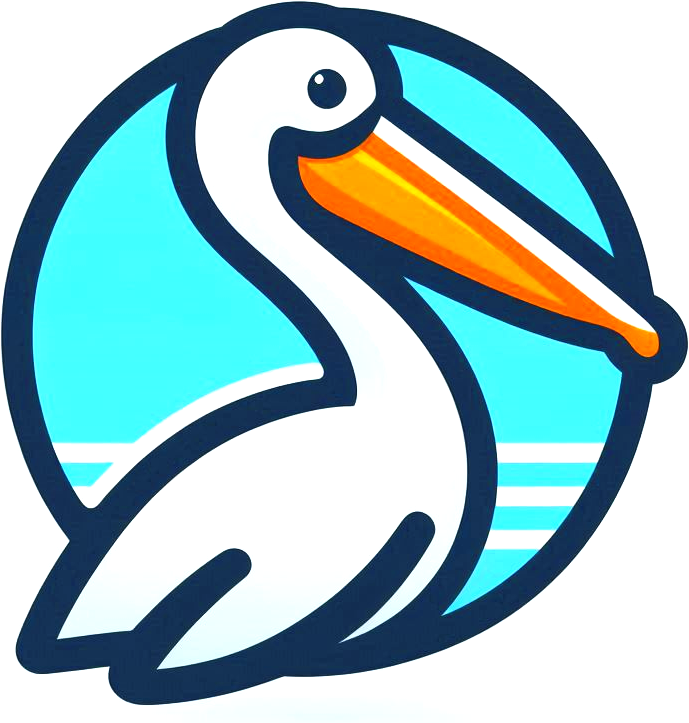} Pelican: Correcting Hallucination in Vision-LLMs via Claim Decomposition and Program of Thought Verification}

\author{
 \textbf{Pritish Sahu\thanks{Equal contribution}},
 \textbf{Karan Sikka\footnotemark[1]},
 \textbf{Ajay Divakaran}\\
 SRI International, Princeton, NJ \\
 \texttt{\{pritish.sahu, karan.sikka, ajay.divakaran\}@sri.com}
 }
\newcommand{\red}[1]{{\textcolor{red}{\text{#1}}}}
\newcommand{\green}[1]{{\textcolor{green}{\text{#1}}}}

\newcommand{\algo}{\texttt{\textbf{Pelican}}\xspace}
\newcommand{\algoV}{\includegraphics[width=0.02\textwidth]{images/pelican.png} \algo\xspace}

\newcommand{\claim}{\mathcal{C}}
\newcommand{\llm}{$V_{L}$}
\newcommand{\mmhal}{MMHal-Bench}
\newcommand{\rellocation}{\texttt{Relative\_location}}

\begin{document}
\maketitle
\begin{abstract}

Large Visual Language Models (LVLMs) struggle with hallucinations in visual instruction following task(s), limiting
 their trustworthiness and real-world applicability.
We propose \algoV-- a novel framework designed to detect and mitigate hallucinations through claim verification.
\algo first decomposes the visual claim into a chain of sub-claims based on first-order predicates. These sub-claims consist of (predicate, question) pairs and can be conceptualized as nodes of a computational graph.
We then use  Program-of-Thought prompting to generate Python code for answering these questions through flexible composition of external tools.
\algo improves over prior work by introducing (1) intermediate variables for precise grounding of object instances, and (2) shared computation for answering the sub-question to enable adaptive corrections and inconsistency identification. We finally use reasoning abilities of LLMs to verify the correctness of the claim by considering the consistency and confidence of the (question, answer) pairs from each sub-claim. 
Our experiments reveal a drop in hallucination rate by $\sim 8\% - 32\%$ across various baseline LVLMs and a $27\%$ drop compared to approaches proposed for hallucination mitigation on MMHal-Bench. Results on two other benchmarks further corroborate our results.

\end{abstract}

\section{Introduction}

Large Vision Language Models (LVLM) have seen significant advancements in recent years \cite{liu2023improved, wu2023multimodal}. 
They typically integrate visual tokens into the embedding space of a Large Language Model (LLM), leveraging the linguistic capabilities of LLMs while incorporating visual information for multimodal understanding. 
Despite substantial performance gains, LVLMs suffer from hallucinations due to limited training data, lack of precise grounding, and over-reliance on language priors \cite{liu2024survey}.


Prior works have focused on scaling training data for reducing hallucinations, as demonstrated by the improved performance of LLaVA-1.5 \cite{liu2023improved} that used many academic datasets during instruction tuning. 
Other works have created high-quality visual instruction tuning datasets. For example, LRV \cite{liu2023mitigating} included diverse examples as well as adversarial questions referencing non-existent objects. Another promising direction has been to improve the model by using variants of reinforcement learning with human feedback (RLHF) to align the model with human preferences \cite{chen2023dress, yu2023rlhf} or by training the model to correct itself through self-feedback during inference \cite{lee2023volcano}. 
Woodpecker \cite{yin2023woodpecker}, inspired by the fact-checking task in NLP \cite{guo2022survey}, recently proposed to correct hallucinations by modeling the problem as verifying and correcting visual claims generated by LVLMs.
The method involved extracting key concepts from the outputs, formulating questions, answering them using visual tools such as VQA, and collating the outputs into a visual claim, which is then used to refine the original output with an LLM. Our work builds upon and extends this claim verification paradigm.

\begin{figure*}[!t]
    \vspace{-3mm}
    \begin{center}
        \includegraphics[width=0.98\linewidth]{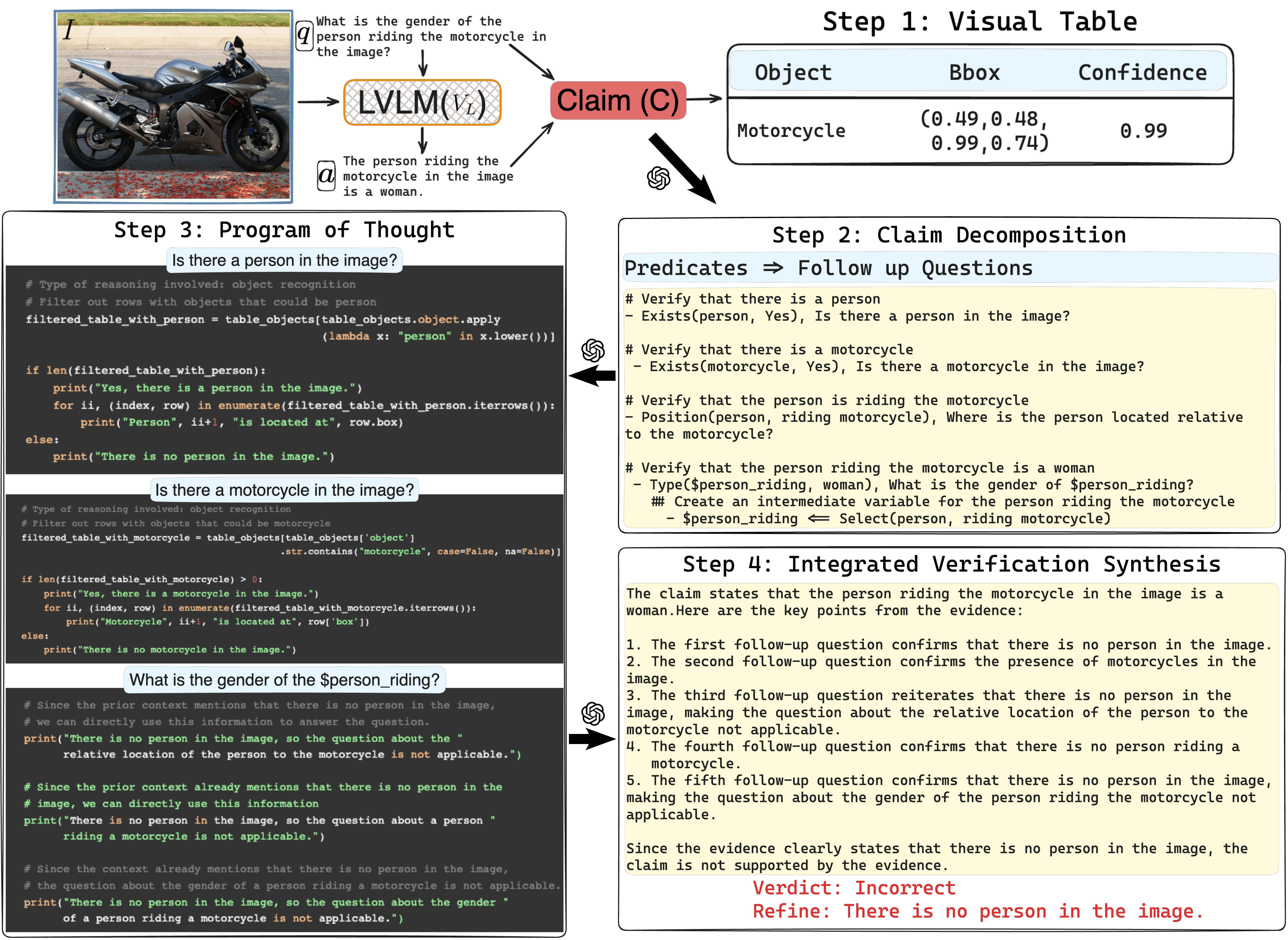}
    \end{center}
    \vspace{-4mm}
    \caption{Overview of \algoV. Given an image ($I$) and a question ($q$), LVLM ($V_L$) outputs an answer ($a$). We transform the pair ($q$, $a$) to our claim ($c$). Our pipeline: Step 1: Visual Table constructs a tabular representation of the image, identifying the locations of visual objects using detection tools. For example, a row is created for the detected ``motorcycle" with its bounding box. Step 2: Claim Decomposition generates a list of granular sub-claims and follow-up questions. This example also shows an intermediate variable, \$person\_riding, that is used for referencing a specific object. Step 3: Program of Thought translates these questions into Python code using the POT approach. Final Step: Integrated Verification Synthesis performs comprehensive reasoning assessments to validate the original claims using the answers from Step 3.}
    \label{fig:model}
    \vspace{-4mm}
\end{figure*}

We propose \algoV (\autoref{fig:model}), a novel and structured pipeline for verifying visual claims to detect and correct hallucinations.
\algo addresses the weaknesses of prior methods for claim verification such as lack of precise grounding, weak contextual integration and visual referencing, and ineffective reasoning over the claim and the visual context.
\algo first breaks down a complex claim into more manageable sub-claims by using a set of predefined first-order predicates tailored to the visual question answering (VQA) task. These sub-claims are represented as a chain of (predicate, question) pairs, with each question stemming from its corresponding predicate.
The resulting chain can be interpreted as a computational graph (\autoref{fig:predicate_graph}), where each node corresponds to a predicate/question. 
The verification of the overall claim is then accomplished by answering the questions in a sequential manner.
\algo uses Program-of-Thought prompting to synthesize Python code that seamlessly integrates external tools with Python operators, offering greater flexibility compared to previous methods like Woodpecker. The introduction of intermediate variables to reference specific object instances is another innovation, which is crucial for precise grounding, particularly in claims involving multiple objects. Furthermore, \algo shares computations from previous nodes in the chain while answering questions, enabling adaptive corrections and the identification of inconsistencies in the reasoning process, distinguishing it from earlier works.
\algo creates a visual table representation, that includes key visual entities, stored as a Pandas dataframe to simplify manipulation of key entities in the code generation step.
The information from the sub-claims is then combined, and the reasoning capabilities of LLM are used to assess the correctness of the claim and generate a refinement in case of hallucinations. Robustness in this step is ensured by using in-context examples and promoting CoT-style reasoning that considers the correctness, confidence, and relevance of the generated answers for each sub-claim. By integrating the flexibility of reasoning in language with the precision of computational methods, \algo is able to achieve strong improvements over SOTA methods. We show significant drop in hallucinations on standard benchmarks (MMHal-Bench, GAVIE) and improve visual understanding accuracy on the MME dataset. Our ablation study reveals the contribution of the key innovations on the final performance. Through qualitative examples, we demonstrate how the model identifies and corrects hallucinated locations.

Our contributions are as follows, we:
\begin{enumerate}
    \item Propose a robust pipeline for identifying hallucination in LVLMs by decomposing the visual claim into sub-claims that consist of questions grounded in first-order predicates. 
    \item Enable precise grounding through intermediate variables for referencing object instances.
    \item Generate Python code to answer sub-questions with external tools, enabling flexible tool composition with Python operators.
    \item Enhance reasoning by sharing computations between questions, allowing for adaptive corrections and identify inconsistencies.
    \item Demonstrate consistent performance improvement over different baseline LVLMs on several benchmarks, as well as improvements relative to existing approaches for mitigating hallucinations. For MMHal-Bench, we reduce  by $\sim 8\% - 32\%$ on LVLMs and $27\%$ over best hallucination mitigation baseline. We also show similar improvements on GAVIE and MME.
    
\end{enumerate}

\section{Related Work}
\label{sec:related}

\paragraph*{Large Visual Language Models and Hallucinations}
Recent LVLMs ~\cite{alayrac2022flamingo,zhu2023minigpt,dai2024instructblip,huang2024language,peng2023kosmos,liu2024visual,liu2023improved} demonstrate superior performance on established benchmarks with strong instruction-following and zero-shot reasoning capabilities. However, these models still suffer from hallucinations and provide incorrect answers or fabricate visual details. This issue arises from several sources: the lack of diverse training data leading to insufficient representation of varied visual contexts~\cite{zhu2023minigpt,dai2024instructblip}, over-reliance on natural language cues \cite{hu2023ciem,liu2024visual,liu2023improved}, a yes-bias tendency to affirmatively answer regardless of the visual content~\cite{liu2024visual}, short text descriptions that do not fully cover the image~\cite{dai2024instructblip}, and synthetic data generation that can introduce non-factual statements~\cite{liu2024visual}. 
We propose a post-hoc procedure to address hallucinations in LVLMs by integrating visual tools with LLMs to ground and reason about their outputs by analyzing a chain of simpler sub-claims.
We refer readers to \autoref{app:related_works} for other works.



\paragraph*{Claim Decomposition}
The use of Language Models (LMs) for fact-checking has gained significant attention. ~\cite{lee2020language} leveraged LLMs to verify simple facts, while ~\cite{atanasova2022fact} addressed the challenge of insufficient evidence. ~\cite{li2023self} enhanced fact-checking by retrieving current information from the web, and ~\cite{cao2023enhancing} incorporated Chain of Thought (CoT) prompting. Subsequent works ~\cite{kotonya2020explainable,atanasova2024generating} focused on generating explanations by summarizing the comments provided by professional fact-checkers. In the domain of fact-checking, claim verification focuses on validating claims by retrieving information from external sources. Recent works, such as ~\cite{li2023self, wang2023explainable}, involve decomposing claims into granular sub-claims for verification with an LLM. Woodpecker \cite{yin2023woodpecker} proposed a similar framework for visual claim verification to correct hallucination. Our work advances these approaches by tightly integrating the chain of sub-claims 
through (1) synthesizing Python code for answering sub-claims, and (2) using shared variables and shared computations between sub-claims to improve efficiency and consistency.


\paragraph*{Tool Calling}
Recent works, such as VisProg, HuggingGPT, and ViperGPT \cite{gupta2023visual, shen2023hugginggpt, schick2023toolformer, suris2023vipergpt} fall under the umbrella of Program-of-Thought prompting \cite{chen2022program} that leverage the strong coding capabilities of LLMs to compose external tools for complex multi-modal tasks.
VisProg uses in-context learning to generate modular programs for visual tasks, while HuggingGPT employs ChatGPT to plan and manage AI models from Hugging Face. 
The proposed work is inspired by these ideas and extends this ability to answer questions via code execution and further
integrate responses from the chain of sub-claims to verify the overall claim.


\section{\algoV}

\paragraph{Problem Formulation:} We are given outputs from a Large Vision Language Model (LVLM) \llm, which provides
an answer $a = \mathcal{V}(I, q)$ for a given image $\mathcal{I} \in \mathcal{R}^{M \times N \times 3}$ (where $M$ and $N$ represent the height and width of the image) and question $q$. We combine the question-answer pair $(q, a)$ into a claim $\claim$~about the image using a Large Language Model (LLM) for the remaining steps.
We formulate the problem of detecting and correcting hallucinations in $a$ as a visual claim verification task \cite{yin2023woodpecker}, which inputs $(\mathcal{I}, \claim)$ and output $(d, r)$, where $d \in \{correct, incorrect\}$ and $r$ refer to the decision regarding the correctness of the claim and a rewrite (if needed) respectively.

We propose \algo (\autoref{fig:model}) for verifying visual claims to detect and correct hallucinations. 
\algo first parses and decomposes the claim into granular sub-claims, consisting of follow-up questions, by leveraging reasoning capabilities of LLMs. Compared to prior works \cite{yin2023woodpecker}, \algo introduces intermediate variables to reference specific instances of objects, enabling precise grounding and visual referencing throughout the verification process.
We then harness the coding capabilities of LLMs to answer the follow-up questions by translating them into Python code using a Program-of-Thought approach \cite{chen2022program} 
This allows for the flexible composition of visual tools using native Python operators. We share the results from previous computations when answering the next question in the chain to facilitate adaptive corrections and catch inconsistencies in the reasoning process often arising from brittleness of visual tools. 
We also handle the limitations of visual tools (e.g., miss-detections) by mapping the image into a tabular representation that consists of visual objects identified in the image.  
We argue that the proposed decomposition implicitly converts the claim into a computational graph with dependencies between sub-questions, while the PoT framework enables \algo to answer questions by combining the flexibility of reasoning in natural language with the precision of computational methods, resulting in a more effective and robust claim verification process.
We next discuss these components in detail.

\subsection{Visual Table} \label{subsec:visual_table}
Our approach relies on tools such as object detectors to ground visual entities in the image. During our initial experiments, we identified several limitations with off-the-shelf object detectors. For instance, closed-vocabulary object detectors are more powerful but cannot detect novel objects, while open-vocabulary detectors may incorrectly identify objects that are not present, leading to false-positives. To address these limitations, we carefully crafted a pipeline (refer to \autoref{app:visual_table} in the appendix) that utilizes YOLO and Grounding-DINO to determine the presence and location of an object in the image. Given the image and the claim, we produce a visual table representation by first parsing the claim using an LLM, with in-context examples, to identify key entities that are tangible and can be visually grounded. This is done to reduce the likelihood of false positives.  
For example, the claim \textit{``The disposable coffee cups are upside down on the nightstand"} will be parsed into $\{cups, nightstand\}$. These entities are then passed to the detectors to create a table $\mathcal{T}$.

Although this step could be offloaded to the code generation step, we chose not to do so for two primary reasons: (1) to avoid complex Python code and allow it to focus on the high-level task of composing tools and reasoning, and (2) because the visual table is stored as a Pandas dataframe, it provides a flexible data structure that can be easily manipulated in Python. To further reduce the false positives from Grounding-DINO, we also utilize the Visual Question Answering (VQA) tool to verify the existence of the detected objects.

\begin{figure}[!t]
    \begin{center}
        \begin{tikzpicture}[
  node distance=4mm and 4mm,
  inner sep=4pt, minimum size=10pt, font=\fontsize{8}{8}\selectfont,
  every node/.style={draw, rectangle, rounded corners, semithick, align=center},
  start node/.style={draw, rectangle, rounded corners, semithick, fill=gray!20, align=center},
  arrow/.style={-Stealth, thick}
]

\node[start node] at (0.0, -0.3) (start) {Exists(dog, Yes)};

\node[right=of start, xshift=-0.8cm, yshift=-0.8cm] (position_right) {Position(dog, right)};
\node[left=of start, xshift=0.8cm, yshift=-0.8cm] (position_left) {Position(dog, left)};

\node[below=of position_right] (select_right) {dog\_right $\leftarrow$ Select(dog, right)};
\node[below=of position_left] (select_left) {dog\_left $\leftarrow$ Select(dog, left)};

\node[below=of select_right] (color_right) {Color(\$dog\_right, brown)};
\node[below=of select_left] (color_left) {Color(\$dog\_left, black)};

\draw[arrow] (start) -- (position_right);
\draw[arrow] (start) -- (position_left);

\draw[arrow] (position_right) -- (select_right);
\draw[arrow] (position_left) -- (select_left);

\draw[arrow] (select_right) -- (color_right);
\draw[arrow] (select_left) -- (color_left);
\end{tikzpicture}
    \end{center}
    \vspace{-3mm}
    \caption{Computational graph representation of the generated sub-claims with predicates as the node and edges defined by their dependencies.}
    \label{fig:predicate_graph}
    \vspace{-3mm}
\end{figure}

\subsection{Claim Decomposition}


Since answering simpler questions about images using external tools (e.g., determining if an object exists) is more reliable, we first propose a novel way to decompose the visual claim $\claim$ into atomic sub-claims. Each sub-claim is represented as a pair consisting of a predicate and its corresponding question $(p_i, q_i)$. 
Each $(q_i)$ can be conceptualized as a node in a computational graph (\autoref{fig:predicate_graph}), where the edges represent the logical connections between the questions. This graph structure will be employed to reason about the claim and determine its veracity by systematically answering the sub-claims.

To achieve this decomposition, we define a set of predicates such as $Exists$ and $Position$, which are loosely structured based on a taxonomy of questions posed in VQA tasks (\autoref{tab:toolbox_details}). These predicates serve as the foundation for deriving specific questions. For instance, a claim involving an object's properties and interactions may be broken down into predicates concerning the presence, attributes, and relations of that object.
We transform the claim $\claim$ into predicates $\mathcal{P}=\{p_i\}_{i=1}^{L}$ by prompting an LLM with in-context examples. We use a set of $\sim10$ examples covering all the predicates and scenarios such as claims with negations. Additionally, we generate a chain of questions $\mathcal{Q}=\{q_i\}_{i=1}^{K}$ grounded in each of the predicates. 

We introduce intermediate variables $v$ to reference specific object instances, which is critical in verifying claims about specific object instances. Moreover, this approach allows us to create dependencies between nodes of the computational graph. This step not only reduces computational redundancy but also improves the reliability of the verification process. 

\subsection{Program of Thought (PoT)-based Sub-Claim Verification}

We now wish to verify each sub-claim, generated in the decomposition step, by generating a visually grounded answer to each sub-question $q_i$. 
While an off-the-shelf instruction-tuned LVLM can be used to answer questions, this approach is constrained by issues in LVLMs, such as hallucinations and lack of interpretability.
Instead, we employ a PoT-based strategy to synthesize programmatic instructions that composes different visual tools to infer the answers to these sub-claims. For example, a question about the color of an object might be translated into a code snippet that first uses a detector to crop the object and then applies a VQA tool to determine the object's color in the image. A key advantage of using code to answer the question is the ability to combine different visual tools flexibly with Python operators.

We denote this step by the function $\lambda$ which inputs the image, current question and the context from prior questions and answer to generate Python code $c_k$ as $\lambda(\mathcal{T}, q_k, \{q_i, a_i\}_{i=1}^{k-1})=c_k$, which is then used to derive the answer as $a_k=\text{exec}(c_k)$, where $\text{exec}$ is a Python interpreter.
We introduce several innovations compared to prior claim verification works to perform robust verification:
\begin{enumerate}
    \item Sharing computations between sub-claims: We provide the answers derived in the previous sub-claims as context when answering the next question in the chain. We found this process to reduce duplicate computations, adaptively correct follow-up questions, and also catch inconsistencies in the reasoning process resulting from errors in the visual tools. For example, a follow-up question about \textit{the color of a car} might not realize that multiple cars are present in the image and may thus end up generating code without loops over the object.
    
    \item Intermediate variable: The intermediate variable created in the decomposition step help the framework to reference specific object instances, e.g., \textit{object on the right}. We found this to improve verification of complex questions requiring contextual reasoning around multiple objects.

    \item Reasoning in Language + Computations: We created our in-context examples to perform reasoning in both language, which provides flexibility, and via computations in the Python code (e.g., through creation of variables and composition with tools).

    \item Visual table: Finally, this module integrates seamlessly with the table prepared in the initial step (\autoref{subsec:visual_table}) of our pipeline. This table, which catalogs the visual entities detected and their attributes, serves as a reference point for validating the existence and properties of objects within the image. By cross-referencing the outputs of the vision tools with the entries in this table, the module can confirm or refute the predicates with a high degree of confidence.
\end{enumerate}

\subsection{Integrated Verification Synthesis}

In this step, we use the answers obtained from the PoT-based verification step to perform a comprehensive reasoning assessment and validate the original claim. This step is crucial for detecting hallucinations and for drawing accurate and reliable conclusions. By aggregating the responses to each sub-claim, the system can make an informed decision about the overall validity of the claim.

We denote this step $d, r = \mathcal{V}(\claim, P, \{q_i, a_i\}_{i=1}^{K})$, where the function $\mathcal{V}$ take in the original claim, predicate, and questions and answers from the sub-claims and outputs $d, r$ which denote the decision and a re-write for claims that contain hallucinations. We realize this with an LLM with a detailed instruction and a few in-context examples that encourages the LLM to verify the claim by considering the evidence along with the consistency and coherence of the generated answers using CoT-style reasoning.
Any discrepancies or inconsistencies are flagged as potential hallucinations, prompting a closer inspection of the questionable elements. This rigorous cross-examination helps in identifying errors that may have arisen due to incorrect or ambiguous interpretations of the visual data.
To mitigate hallucinations, we use the same LLM to rewrite the original claim based on the verified answers. This involves rephrasing the claim to eliminate elements that were identified as hallucinated or incorrect. 

Through this integrated verification synthesis, our pipeline not only verifies the accuracy of claims but also actively improves the clarity and correctness of the information presented. This approach significantly reduces the risk of hallucinations, ensuring trustworthiness and precision. 

\section{Experiment}

\subsection{Experimental Setup}
\paragraph{Tasks and Benchmarks.} We use the following LVLM benchmarks for evaluation.

\begin{enumerate}
    \item \textbf{\mmhal}~\cite{sun2023aligning} evaluates informativeness and hallucinations in responses by comparing model outputs with human responses and object labels using GPT-4. The benchmark includes 96 image-question pairs across 8 categories and 12 topics, focusing on open-ended questions. We report the informativeness score and hallucination rate.
    \item \textbf{GAVIE}~\cite{liu2023mitigating} evaluates hallucination by measuring the accuracy and relevancy (instruction-following ability) of responses in an open-ended manner without needing human-annotated ground truth answers
    We selected a random subset of 250 questions and use GPT-4o for evaluation. 
    \item \textbf{MME}~\cite{fu2023mme} evaluates LVLMs' perception and cognition through ``Yes" or ``No" questions for each test image. It has 14 subtasks: 10 for perception and 4 for cognition. Following ~\cite{yin2023woodpecker}, we use the existence, count, position, and color subsets to assess hallucination at object and attribute levels. We report the sum of accuracy and accuracy plus (when the model correctly answers both ``Yes" and ``No" questions per image).
\end{enumerate}

We discuss evaluation on hallucination benchmarks (\mmhal, GAVIE) in \autoref{sec:halcorrect}, and on the visual understanding benchmark (MME) in \autoref{sec:vis_understanding}.

\begin{table*}[!h]
     \centering
     \begin{adjustbox}{width=\linewidth}
     \begin{tabular}{lccccccccccc}
     \toprule 
     \multirow{2}{*}{\textbf{Model}} & \multirow{2}{*}{\algoV} & \multicolumn{2}{c}{\textbf{\mmhal}} &  \multicolumn{3}{c}{\textbf{GAVIE}} & \multicolumn{5}{c}{\textbf{MME}} \\ \cmidrule(l){3-4} \cmidrule(l){5-7}  \cmidrule(l){8-12}   \addlinespace[-0.03cm]
      &  & \multicolumn{1}{c}{Score $\uparrow$} & Hal-Rate $\downarrow$ &  \multicolumn{1}{c}{Acc $\uparrow$} & \multicolumn{1}{c}{Rel $\uparrow$} & \multicolumn{1}{c}{Avg $\uparrow$}  & \multicolumn{1}{c}{Existence} & \multicolumn{1}{c}{Count} & \multicolumn{1}{c}{Position} & \multicolumn{1}{c}{Color} & \multicolumn{1}{c}{Total} \\ \midrule 
     \multirow{2}{*}{InstructBlip-7B} & $\times$ & $1.71$ & $0.66$ &  $5.61$ & $7.1$ & $6.36$ & $\textbf{185}$ & $58$ & $58$ & $143$ & $444$ \\ \addlinespace[-0.03cm]
                                      & $\checkmark$ & $\textbf{2.26}$ & $\textbf{0.51}$ &   $\textbf{6.66}$ & $\textbf{7.6}$ & $\textbf{7.13}$ & $175$ & $\textbf{153}$ & $\textbf{147}$ & $\textbf{152}$ & $\textbf{627}$ \\ \addlinespace[-0.03cm]
     \multirow{2}{*}{mPlug-OWL} & $\times$ & $1.34$ & $0.74$ &   $3.88$ & $7.1$ & $5.49$ & $95$ & $48$ & $50$ & $55$ & $248$ \\ \addlinespace[-0.03cm]
                                  & $\checkmark$ & $\textbf{2.35}$ & $\textbf{0.50}$ &   $\textbf{6.55}$ & $\textbf{7.5}$ & $\textbf{7.03}$ & $\textbf{175}$ & $\textbf{150}$ & $\textbf{133}$ & $\textbf{153}$ & $\textbf{611}$ \\ \addlinespace[-0.03cm]
     \multirow{2}{*}{LLaVA-v1.5-7B} & $\times$ & $2.02$ & $0.61$ &   $5.64$ & $7.4$ & $6.52$ & $\textbf{175}$ & $88$ & $103$ & $105$ & $471$ \\ \addlinespace[-0.03cm]
                                   & $\checkmark$ & $\textbf{2.27}$ & $\textbf{0.52}$ &   $\textbf{6.51}$ & $\textbf{7.6}$ & $\textbf{7.06}$ & $\textbf{175}$ & $\textbf{153}$ & $\textbf{122}$ & $\textbf{140}$ & $\textbf{590}$ \\ \addlinespace[-0.03cm]
     \multirow{2}{*}{mPlug-OWL2} & $\times$ & $1.88$ & $0.65$ &   $5.82$ & $\textbf{7.9}$ & $6.86$ & $150$ & $83$ & $58$ & $118$ & $409$ \\ \addlinespace[-0.03cm]
                                   & $\checkmark$ & $\textbf{2.26}$ & $\textbf{0.51}$ &   $\textbf{6.42}$ & $7.6$ & $\textbf{7.01}$ & $\textbf{175}$ & $\textbf{142}$ & $\textbf{121}$ & $\textbf{140}$ & $\textbf{578}$ \\ \addlinespace[-0.03cm]
     \multirow{2}{*}{LLaVA-v1.6-7B} & $\times$ & $\textbf{3.24}$ & $0.41$ &   $6.13$ & $\textbf{7.8}$ & $\textbf{6.97}$ & $\textbf{195}$ & $155$ & $138$ & $\textbf{190}$ & $678$ \\ \addlinespace[-0.03cm]
                                             & $\checkmark$ & $3.04$ & $\textbf{0.38}$ &   $\textbf{6.33}$ & $7.5$ & $6.92$ & $185$ & $\textbf{172}$ & $\textbf{178}$ & $180$ & $\textbf{715}$ \\ \bottomrule
     \end{tabular}
     \end{adjustbox}
     \vspace{-2mm}
     \caption{\hspace*{0mm}Results on hallucination (\mmhal, GAVIE) and visual understanding (MME) benchmarks. \mmhal scores range from 0-6, with hallucination rate (Hal-Rate) indicating the proportion of scores below 3. GAVIE measures accuracy (Acc) and relevancy (Rel) on a 0-10 scale, with Avg representing their average. MME reports the sum of accuracy and accuracy plus for each category (object-level and attribute-level), with Total representing the sum across all categories. $\checkmark$ denotes results corrected by \algoV. The best result for each setting is highlighted in bold.}
     \label{tab:hal_mme_results}
      
     \begin{adjustbox}{width=\linewidth}
     \begin{tabular}{lcccccccccc}
          \toprule
			\multirow{2}{*}{\textbf{Model}} & \multicolumn{2}{c}{\textbf{\mmhal}} &  \multicolumn{3}{c}{\textbf{GAVIE}} & \multicolumn{5}{c}{\textbf{MME}} \\ \cmidrule(l){2-3}  \cmidrule(l){4-6} \cmidrule(l){7-11}   \addlinespace[-0.03cm]
      		& \multicolumn{1}{c}{Score $\uparrow$} & Hal-Rate $\downarrow$  & \multicolumn{1}{c}{Acc $\uparrow$} & \multicolumn{1}{c}{Rel $\uparrow$} & \multicolumn{1}{c}{Avg $\uparrow$}  & \multicolumn{1}{c}{Existence} & \multicolumn{1}{c}{Count} & \multicolumn{1}{c}{Position} & \multicolumn{1}{c}{Color} & \multicolumn{1}{c}{Total} \\ \midrule 
            Woodpecker (InstructBlip)\ & $1.71$ & $0.67$ &   $5.48$ & $7.4$ & $6.44$ & $160$ & $78$ & $90$ & $100$ & $428$ \\
            Woodpecker (LLaVA-v1.6-7B)\ & $1.73$ & $0.66$ &  $5.38$ & $7.4$ & $6.39$ & $165$ & $78$ & $90$ & $100$ & $433$ \\
          	Volcano-7B\ & $2.21$ & $0.57$ &   $5.32$  & $7.5$ & $6.41$ & $195$ & $152$ & $107$ & $160$ & $614$ \\
          	Volcano-13B\ & $\underline{2.44}$ & $\underline{0.52}$ &  $\underline{5.97}$ & $\textbf{8.1}$ & $\textbf{7.04}$ & $\textbf{195}$ & $\underline{158}$ & $\underline{118}$ & $\textbf{185}$ & $\underline{656}$ \\
          	\algoV & $\textbf{3.04}$  & $\textbf{0.38}$ &   $\textbf{6.33}$ & $\underline{7.5}$  & $\underline{6.92}$ & $\underline{185}$ & $\textbf{172}$ & $\textbf{178}$ & $\underline{180}$ & $\textbf{715}$ \\ \bottomrule
     \end{tabular}
     \end{adjustbox}
     \vspace{-2mm}
     \caption{\hspace*{0mm}Results compared against Woodpecker \cite{yin2023woodpecker} and Volcano \cite{lee2023volcano}, two methods previously proposed for correcting hallucination. The best scores are highlighted in bold, and the second-best scores are underlined.}
     \label{tab:hal_mme_us_vs_baseline_results}
     \vspace{-4mm}
\end{table*}

\paragraph{Baselines.} We use InstructBlip~\cite{dai2024instructblip}, LLaVA-v1.5-7B~\cite{liu2023improved}, LLaVA-v1.6-7B~\cite{liu2024llavanext}, mPlug-OWL~\cite{ye2023mplug}, and mPlug-OWL2~\cite{ye2023mplug2} as the baseline models whose responses will be evaluated with \algo. 
We also compare with approaches proposed to address hallucinations-- Woodpecker ~\cite{yin2023woodpecker} and Volcano ~\cite{lee2023volcano}. Refer to \autoref{app:implementation} for implementation details.



\subsection{Hallucination Detection and Mitigation}
\label{sec:halcorrect}

\autoref{tab:hal_mme_results} and \autoref{tab:hal_mme_us_vs_baseline_results} show the performance evaluation of \algo on the hallucination benchmarks-- \mmhal~ and GAVIE.
The results highlight the superiority of \algo in both reducing hallucinations and improving performance on the given question-answering task.
\autoref{tab:hal_mme_results} demonstrates improvements by using \algo to refine the responses of several LVLMs.
For each row, $\checkmark$ and $\times$ denote performance with and without applying \algo respectively. 
Across all three hallucination benchmarks, \algo improves the relevancy and accuracy scores and reduces the hallucination rate. This gain is particular larger for earlier models such as InstructBlip, mPlug-OWL, LLaVA-v1.5 and mPlug-OWL2. We observe the hallucination rate to drop by $\sim15-32\%$ since \algo is able to address issues such as weak visual grounding (by using external tools) and the bias towards ``Yes" answers.
LLaVA-v1.6 is a recent LVLM that leverages dynamic high-resolution and high-quality data to achieve SOTA performance. 
We achieve a reduction in hallucination score on \mmhal~of $0.3$ with \algo. Since we use LLaVA-v1.6 as the VQA tool in \algo, this result shows that our approach does not completely rely on this tool and composes different tools to mitigate hallucinations (refer to \autoref{tab:hal_mme_us_vs_baseline_results} for results with a different VQA tool).
Our algorithm obtains a lower relevancy score on GAVIE for both LLaVA-v1.6 and mPlug-OWL2 since relevancy focuses on instruction following performance. 
This occurs since the refinement generated by \algo is conditioned only on the original claim and may not directly respond to the original question due to lost context. 

\begin{table*}[!h]
     \centering
     \begin{adjustbox}{width=\linewidth}
     \begin{tabular}{lccccccc}
     \toprule 
     \multirow{2}{*}{\textbf{Model}} & \multicolumn{2}{c}{\textbf{MMHal-Bench}} & \multicolumn{5}{c}{\textbf{MME}} \\ \cmidrule(l){2-3} \cmidrule(l){4-8} \addlinespace[-0.03cm]
      
      & \multicolumn{1}{c}{Score $\uparrow$} & Hal-Rate $\downarrow$ & \multicolumn{1}{c}{Existence} & \multicolumn{1}{c}{Count} & \multicolumn{1}{c}{Position} & \multicolumn{1}{c}{Color} & \multicolumn{1}{c}{Total} \\ \midrule
     \algo  & $\textbf{2.27}$ & $\textbf{0.52}$ & $\textbf{175}$  & $\textbf{153}$ & $\textbf{122}$ & $140$ & $\textbf{590}$ \\ \addlinespace[-0.03cm]
     \algo w/o sh\_var ($v$) & $2.23$ & $0.55$ & $170$ & $148$ & $98$ & $\textbf{160}$ & $576$  \\ \addlinespace[-0.03cm]
     \algo w/o sh\_var($v$), sh\_comp & $2.24$ & $0.54$ & $170$ & $148$ & $97$ & $70$ & $485$  \\ \addlinespace[-0.03cm]
     \algo w VQA(LLaVA-v1.5) & $2.20$ & $0.55$ & $170$ & $143$ & $123$ & $145$ & $581$  \\ \bottomrule
     \end{tabular}
     \end{adjustbox}
     \vspace{-3mm}
     \caption{Ablation study to highlight the performance contribution of different innovations proposed in \algoV. ``sh\_var" refers to shared variable and ``sh\_comp" refers to shared computations.}
     \label{tab:hal_mme_ablation_results}
\end{table*}

\autoref{tab:hal_mme_us_vs_baseline_results} compares \algo against two methods designed to reduce hallucinations: Woodpecker, which uses a claim verification pipeline, and Volcano, which employs a self-feedback guided refinement model.
For a fair comparison with Woodpecker, we ran the author's implementation with both InstructBlip and LLaVA-v1.6 as the VQA tool. 
Our model shows a $27\%$ drop in hallucination from the best-performing baseline (Volcano-13B), highlighting the robustness and reliability of \algo in handling multimodal hallucinations. Moreover, the $38\%$ reduction in hallucination rate compared to Woodpecker emphasizes our algorithmic innovations in visual claim verification (Section 1).

\begin{figure*}[!ht]
    \vspace{-1mm}
    \begin{center}
        \includegraphics[width=0.98\linewidth]{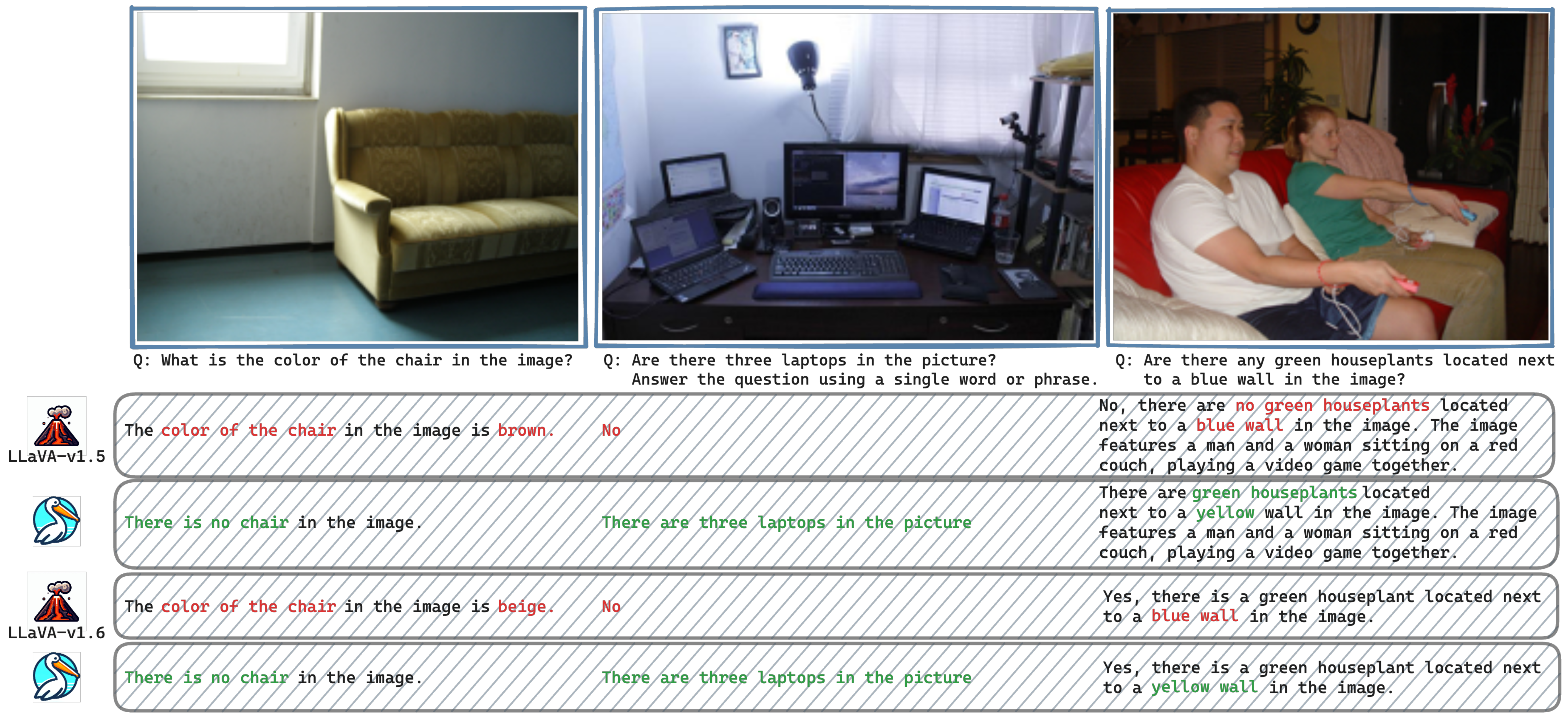}
    \end{center}
    \vspace{-3mm}
    \caption{An illustration of hallucination in LVLMs. Three examples showcasing different types of question-answering styles, where both LLaVA-v1.5 and LLaVA-v1.6 \red{hallucinates} to the question. \algoV refines the answer from these models exhibiting significantly \green{reduced hallucinations}.} 
    \label{fig:demo}
    \vspace{-3mm}
\end{figure*}

\subsection{Visual Understanding}
\label{sec:vis_understanding}

We evaluate \algo on visual understanding using the MME benchmark, as shown in the last column of \autoref{tab:hal_mme_results} and \autoref{tab:hal_mme_us_vs_baseline_results}. We focus on four key categories that significantly contribute to hallucination: object-level (existence and count) and attribute-level (position and color). Our results demonstrate that \algo significantly improves visual understanding when integrated with these models. For instance, mPlug-OWL initially underperforms, but when integrated with our approach, we observe a remarkable $146\%$ improvement. Moreover, compared to LLaVA-v1.6, which exhibits superior performance, we achieve an overall $5.4\%$ improvement, particularly over $\sim 40\%$ improvement in the count and position categories, which are considered the most challenging.
To accurately answer questions about count and position, the model must localize, and reference regions based on size, considering that objects can be in the foreground or background and may blend with similar colors from the surroundings. Furthermore, the model needs to contextualize the relative locations of objects within the image.
Overcoming these challenges requires the model to perform detailed reasoning about the key elements and their attributes and relationships.
Similar patterns are observed in \autoref{tab:hal_mme_us_vs_baseline_results}, where we outperform all baseline models, especially in count and position. However, in some cases, Volcano-13B achieves higher scores ($10\%$ and $5\%$ higher accuracy in existence and color) compared to \algo. This can be attributed to Volcano-13B being trained specifically to rewrite its response based on self-critique and using a larger LLM backbone.

\subsection{Ablation Studies}
\label{sec-ablation}

\autoref{tab:hal_mme_ablation_results} shows our ablation study conducted on \algo to 
to assess the impact of different components on performance. 
The study includes four configurations: original (\algo), without shared variables between sub-claims (\algo sh\_var($v$)), without shared variables and shared computations (\algo w/o sh\_var($v$), sh\_comp), and lastly we replace LLaVA-v1.6 with LLaVA-v1.5 for VQA tasks (``\algo w VQA"). We evaluated these configurations on \mmhal~ and MME benchmarks. The results demonstrate that removing shared variables and shared computations between sub-claims leads to a noticeable fall in performance, particularly on the MME benchmark, where the total score drops from $590$ to $485$. In particular, the drop on position is higher ($122 \to 98$) as the shared variables are responsible for referencing specific object instances which is important for such question (e.g., \textit{Is the motorcycle on the right side of the bus?}). 
Additionally, replacing LLaVA-v1.6 with LLaVA-v1.5 for VQA tasks results in reduction in the score and a small increase in hallucination rate on MMHal-Bench and a $9$ point decrease in Total on MME. This shows both that \algo is robust to the underlying VQA tools, but overall performance will improve with better tools. \autoref{fig:demo} presents a qualitative comparison with LLaVA-v1.5 and LLaVA-v1.6, illustrating that these models struggle to accurately detect object existence, color, location, and count, leading to overall failure. In contrast, our proposed approach successfully addresses these challenges.

Our performance improvement primarily stems from two key innovations: shared computations and shared variables. Additionally, the claim decomposition process does not involve visual information, making it agnostic to whether LVLMs or LLMs are used, as both models operate solely on textual content. Initial experiments with a REACT agent, using a single LLM for sequential tasks, failed due to brittleness in long-range reasoning and an inability to adapt to the varied nature of code generation. Overall, these findings highlight the effectiveness of our proposed approach and the critical role of each component in reducing hallucinations and enhancing visual understanding.

\section{Conclusion}

We propose \algoV, a novel solution for detecting and mitigating hallucinations via visual claim decomposition and program of thought. \algo is a structured pipeline that breaks down claims into granular sub-claims and uses a robust verification mechanism to ensure thorough checking and validation of each aspect of the visual information. Our approach improves upon the state-of-the-art benchmarks results on hallucination and visual understanding. We also show improved performance on visual tasks  where models often struggle and tend to hallucinate, highlighting \algo's strength in providing comprehensive visual understanding. Our unique principled approach involves decomposing claims into sub-claims, using shared variables, and leveraging in-context examples to guide Python code execution for precise and deterministic answers. Moreover, we employ an LLM to reason over these answers, mimicking human-like comprehension and argumentation. The process  systematically addresses and corrects hallucinations, ensuring higher accuracy and reliability in both visual and textual understanding, making it a valuable addition to any LVLM.
\newpage
\section{Limitations} \label{sec:limitation}
We discuss the limitations of \algo in this section.
\begin{enumerate}
    \item \textbf{Brittleness of Visual Tools:} We used visual tools such as VQA and object detection models to verify individual sub-claims, but these tools often fail. For instance, YOLO and DETR object detectors frequently struggle with (1) very small objects, (2) objects out of their normal context, and (3) visual entities outside the detector's vocabulary. Although Grounding-DINO can detect novel objects, it tends to produce several false positives. Similar failures were observed with the LLaVA-v1.6 VQA tool, which often failed to correctly answer questions regarding object attributes. Despite our efforts to mitigate these limitations through the visual table representation and object class verification, some failures persist.
    \item \textbf{Randomness and Lack of Consistency in LLM outputs}:  Even with the temperature set to 0, we observed randomness in the outputs of our pipeline. Consistently extracting visual entities in the key concept extraction step was particularly challenging. For example, the LLM often broke compound nouns such as ``sports ball" into ``ball" or ``bath towel" to ``towel" causing the object detector to fail. When the LLM reduces ``bath towel" to ``towel," YOLO/DETR fail to detect ``towel" because it is not in their list of classes. Even Grounding-DINO, which can detect ``bath towel" fails to detect ``towel."
    The sensitivity of outputs to the prompt was manageable in claim decomposition and code generation, but the code generation step sometimes produced code that would fail in the Python interpreter. We addressed this issue by generating the code three times until it did not produce any errors.
    \item \textbf{Issues with using a claim for verification:} Current pipeline transforms the (question, answer) pair into a claim, which is used for verification. The refined output produced by our pipeline is directly used for evaluation, instead of transforming it back into an answer conditioned on the context. This often results in lower relevancy scores, which measure instruction-following performance.
    \item \textbf{Ability to handle conflicting evidence:} In specific cases, we observe our pipeline unable to handle conflicting evidence, usually due to tool failures. In such cases, the pipeline may incorrectly declare the claim as true or false, leading to inaccuracies.
\end{enumerate}

\section{Ethical Considerations}
Detecting and mitigating hallucinations inherently enhances ethical soundness. Our claim-decomposition  fosters transparency and empowering enhanced control over ethical considerations. It also fosters broader sets of checks and balances that render ethical abuses more challenging, thus reducing the potential for misuse.

\bibliography{acl_latex}

\appendix
\newpage
\section{Predicates and Tools} \label{sec:toolbox_details}
We list out the predicates and tools invoked for each predicates in \autoref{tab:toolbox_details}. We defined these predicates by using a taxonomy of question-types from prior works on VQA \cite{antol2015vqa} and multimodal hallucinations \cite{bai2024hallucination}.

\begin{table*}[h]
    \centering
    \begin{adjustbox}{width=\linewidth}
    \begin{tabular}{ccc}
        \toprule
        \textbf{Predicates} & \textbf{Tools} & \textbf{Reasoning} \\ \midrule
        Exist & Object Detector & Trained specifically to detect and locate objects \\ 
        OCR & VQA & Trained on OCR-based datasets \\ 
        Count & Object Detector & Detecting every instance of object, assists in counting. \\ 
        Attribute & VQA & Trained on paired image, caption datasets such as COCO, Visual Genome \\ 
        Location (rel) & VQA, \rellocation & Bbox from object detectors provide relative location \\
        Scene & VQA & Trained on caption datasets \\ \bottomrule
    \end{tabular}
    \end{adjustbox}
    \caption{Different predicates along with tools and the type of reasoning involved. \rellocation is a function to determine the relative location of an object relative to another object based on their bounding boxes.}
    \label{tab:toolbox_details}
\end{table*}

\section{Visual Table}
\label{app:visual_table}
In \autoref{subsec:visual_table}, we propose a pipeline to detect objects referenced in a claim by integrating both YOLOv9 and Grounding DINO. Here, we provide a detailed overview of the pipeline that takes the best of both worlds, YOLOv9 which is a closed-vocabulary set and Grounding DINO which offers an open-vocabulary set, allowing us to balance the strengths and weaknesses of both tools. Closed-vocabulary models like YOLOv9 are reliable for the classes they are trained on but fail to detect novel classes. Conversely, open-vocabulary models like Grounding DINO can detect novel objects but may produce false positives, such as mistakenly identifying a pillow that isn't present.

To address these issues, our pipeline first checks if the object class is present in YOLOv9's closed set. If the class exists and YOLOv9 does not detect the object, we conclude the object is not found. If the class is not present, we use Grounding DINO to verify the object's presence. Additionally, for the GAVIE benchmark, we incorporate a bounding box verification tool. This tool uses the coordinates of detected objects and queries a VQA module (LLaVA-v1.6) to confirm if the objects are indeed present in the image, enhancing the accuracy of our detection system.

\section{Instruction \& Prompts Templates}
\label{sec:prompt}
In this section, we provide the list of prompt templates used in various steps of \algo.

We put the instructions and prompts we use for extracting visual elements, claim decomposition, program of thought verification and integrating verification. 

\begin{figure*}
    \centering
    \includegraphics[width=\textwidth]{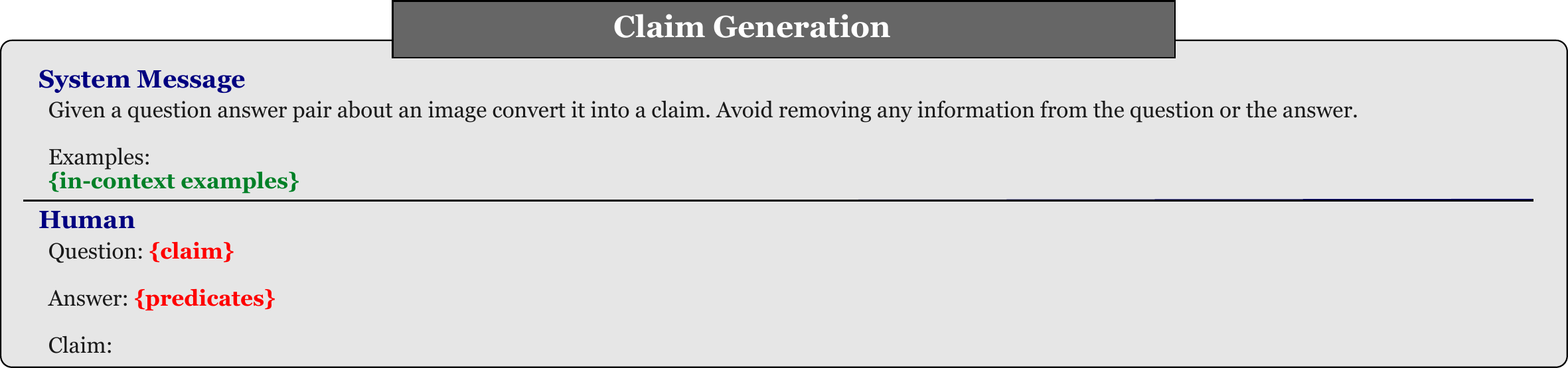}
    \label{fig:claim_generation}
    \caption{Prompt template to generate a claim using both the question and answer.}
\end{figure*}

\begin{figure*}
    \centering
    \includegraphics[width=\textwidth]{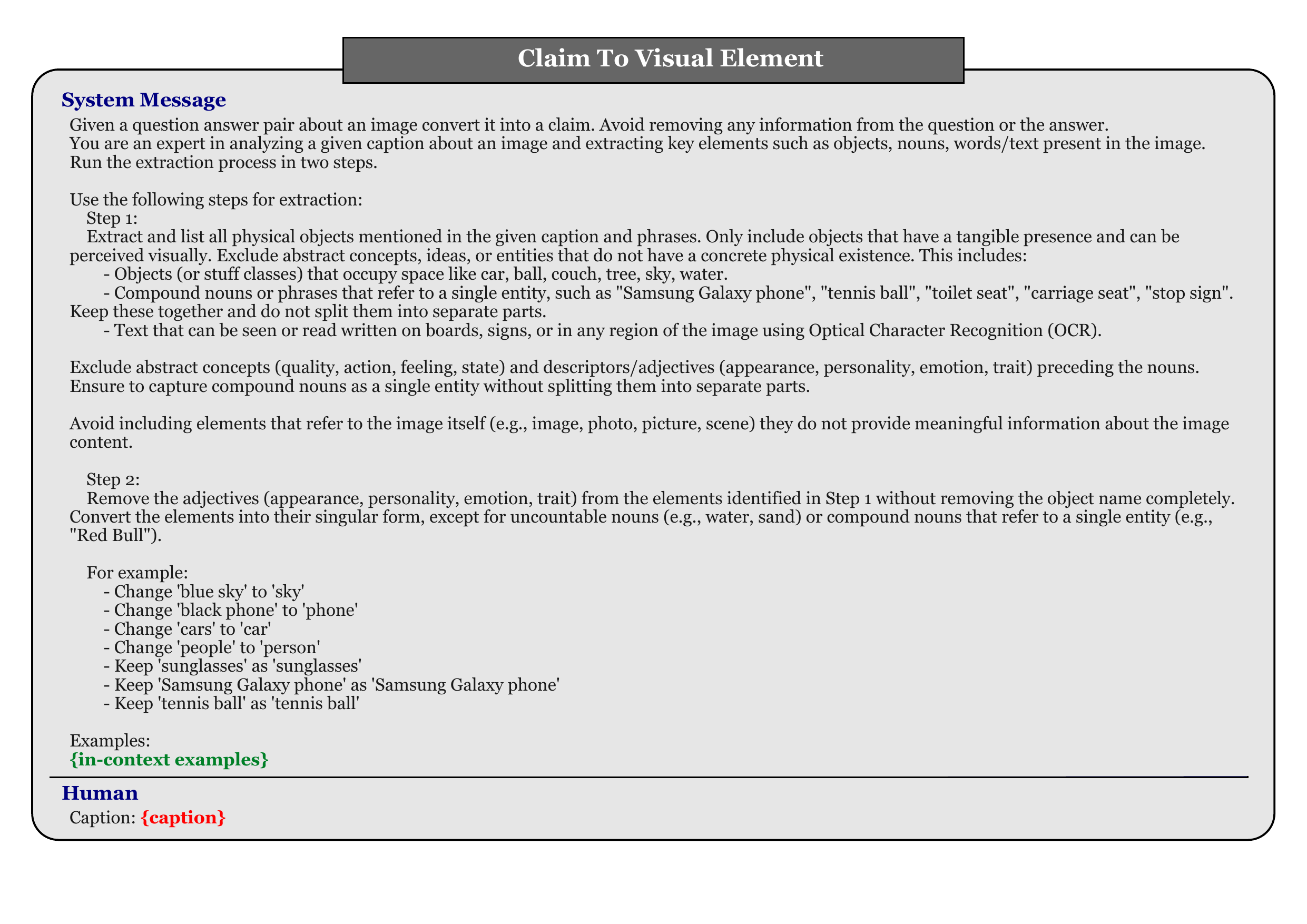}
    \label{fig:claim2objects}
    \caption{Template for prompting LLM to perform key concept extraction.}
\end{figure*}

\begin{figure*}
    \centering
    \includegraphics[width=\textwidth]{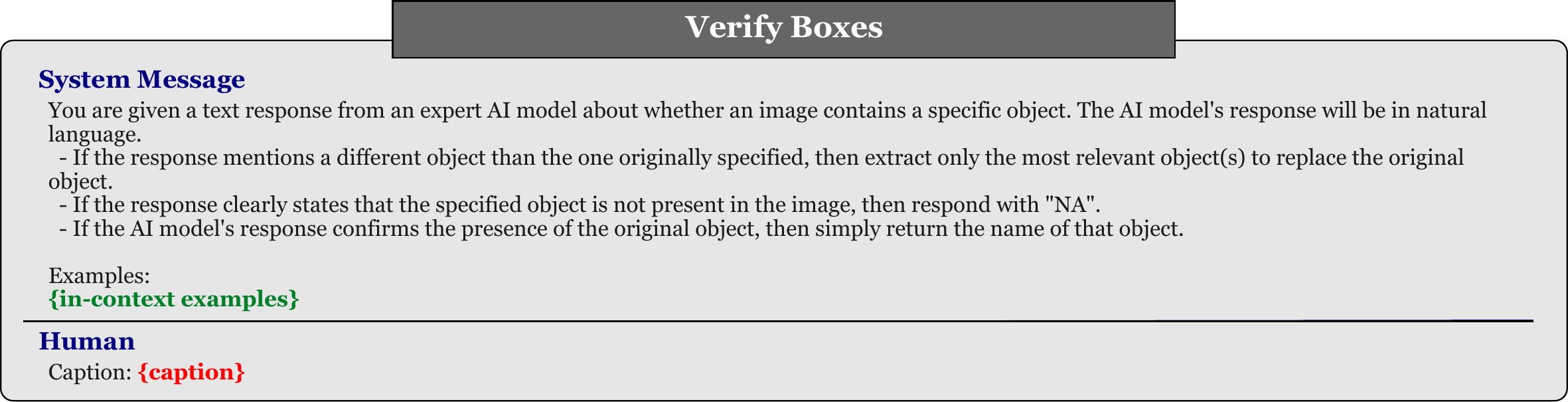}
    \label{fig:verify_boxes}
    \caption{Prompt template to detect the existence of an object in a boxed area.}
\end{figure*}

\begin{figure*}
    \centering
    \includegraphics[width=\textwidth]{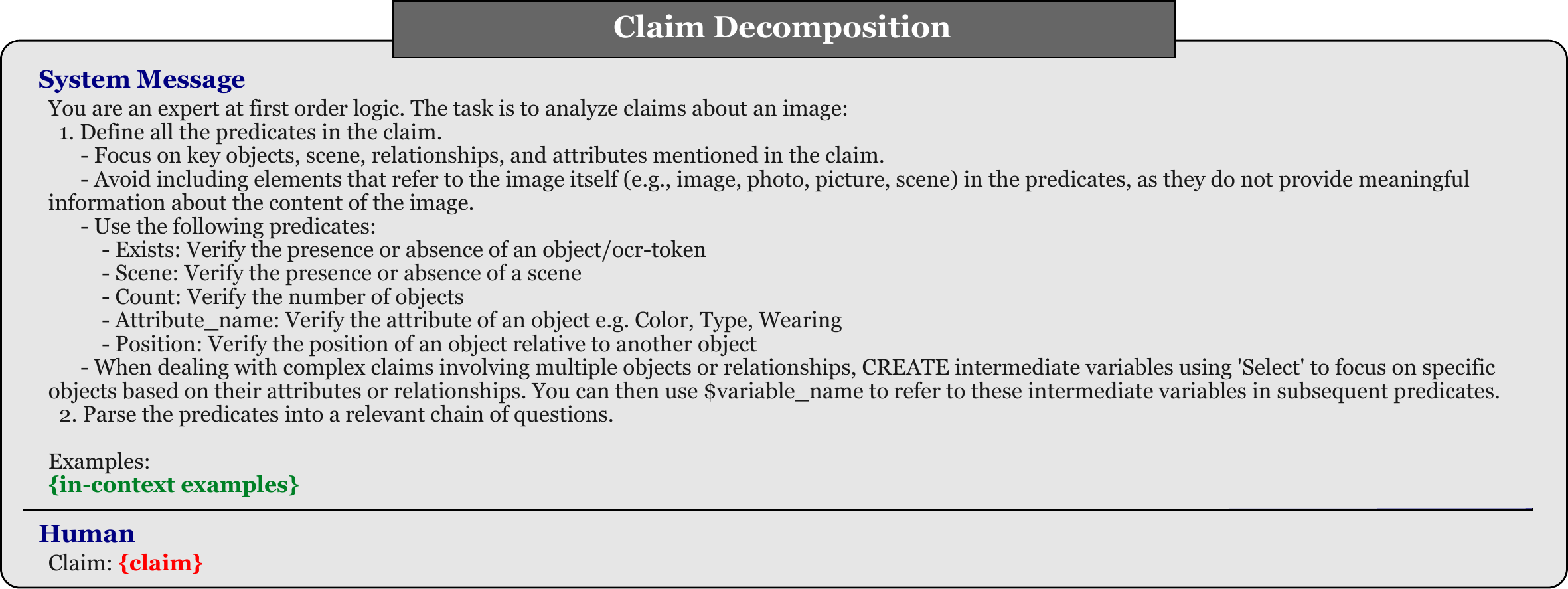}
    \label{fig:claim_decompose}
    \caption{Template for prompting LLM to decompose the claim into a list of predicates.}
\end{figure*}

\begin{figure*}
    \centering
    \includegraphics[width=\textwidth]{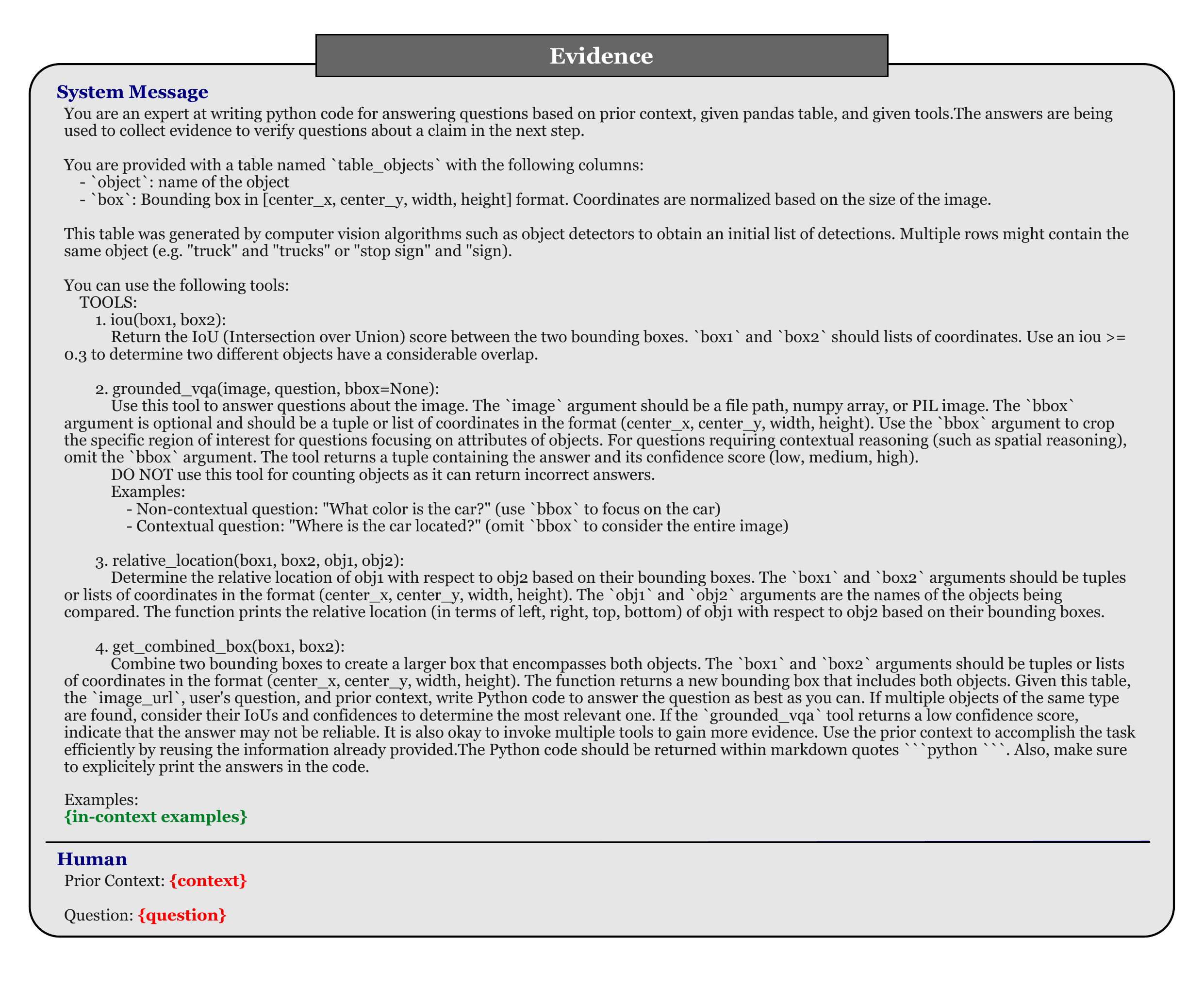}
    \label{fig:evidence}
    \caption{Prompt Template for function calling based on the task type for, e.g., to answer about visual attributes it calls grounded\_vqa.}
\end{figure*}


\begin{figure*}
        \centering
        \includegraphics[width=\textwidth]{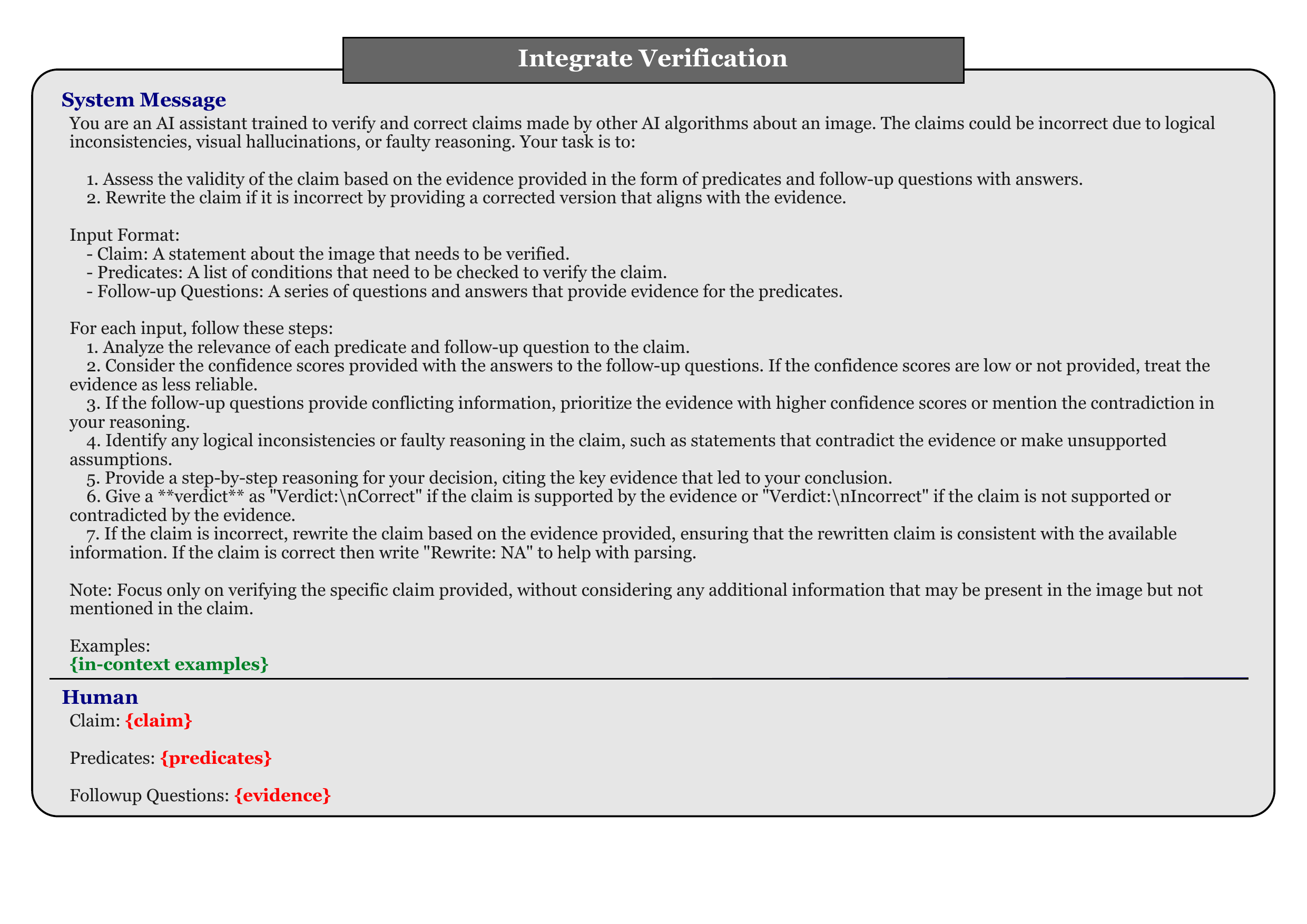}
        \label{fig:prompt_verification_and_rewrite}
        \caption{Prompt template to summarize the sub-claim responses and verify whether the claim was correct. Finally, rewrite that claim (if it has hallucinations) based on the responses. }
\end{figure*}

\section{Additional Related Works}
\label{app:related_works}
Apart from the works presented in \autoref{sec:related} we discuss a few other relevant works.

\paragraph*{Large Visual Language Models and Hallucinations} Several approaches have been proposed to mitigate hallucinations in Large Visual Language Models (LVLMs). One line of work employs variants of Reinforcement Learning from Human Feedback (RLHF) to enhance LVLMs ~\cite{yu2023rlhf, chen2023dress, jing2024fgaif}. DRESS~\cite{chen2023dress}, for instance, utilizes fine-grained feedback in the form of critique and refinement obtained from Language Models (LLMs) to align the LVLM with human preferences. On the other hand, RLHF-V acquires segment-level feedback directly from humans and applies dense direct preference optimization to achieve alignment. A recent work, Volcano \cite{lee2023volcano}, proposes distilling the capability of self-critique within the LVLM to improve performance during inference. This is accomplished by generating instruction-following data containing critique and improved responses from LLMs, which is then used to fine-tune the LLaVa-1.5 model.

\section{Implementation Details:}
\label{app:implementation}
\paragraph{\algo} Our pipeline uses specific prompts, listed in \autoref{sec:prompt}, for different components of our pipeline. 
These prompts were fed to GPT-4o for producing responses. 
For accurate understanding of visual content, we use several tools as described in \autoref{sec:toolbox_details}. Specifically, we realize these tools using:

\begin{itemize}
    \item \textbf{Object Detection:} We used Yolo-v9~\cite{wang2024yolov9} or DETR~\cite{carion2020end} based on benchmark to localize image regions specific to entities mentioned in the claim. Grounding Dino was used to detect objects not present in the vocabularly of these detectors.
    \item \textbf{VQA:} We used LLaVA-v1.6 for visual question answering the attribute related questions related to the image.
    \item \rellocation: We implemented this function in python to determine the relative location (left, right, top, bottom) of object1 with respect to object2 by comparing their bounding boxes. 
\end{itemize}
We also use Pandas library for efficient data manipulation and analysis of the visual table. We would like to note that we guided the LLM generating Python code to use these tools with a few in-context examples. In this case, we often found the LLM to compose different combinations than what we had provided in these examples.


\paragraph{Baselines:}
For our experiments, we ran the inference for all the models locally with temperature set to 0.
\begin{itemize}
    \item InstructBlip~\cite{dai2024instructblip}: We used the model ``instructblip-vicuna-7b"\footnote{https://huggingface.co/Salesforce/instructblip-vicuna-7b} provided by authors on Huggingface. 
    \item LLaVA-v1.5-7B~\cite{liu2023improved} We used the code and the model provided\footnote{https://github.com/haotian-liu/LLaVA} by the authors.
    \item LLaVA-v1.6-7B~\cite{liu2024llavanext}: We used the model ``llava-v1.6-mistral-7b-hf"\footnote{https://huggingface.co/llava-hf/llava-v1.6-mistral-7b-hf} provided by the authors on Huggingface.
    \item mPlug-OWL~\cite{ye2023mplug}: We use the model and the code provided\footnote{https://github.com/X-PLUG/mPLUG-Owl} by the authors.
    \item mPlug-OWl2~\cite{ye2023mplug2}:We use the model and the code provided\footnote{https://github.com/X-PLUG/mPLUG-Owl} by the authors.
    \item Volcano~\cite{lee2023volcano}: We used the code\footnote{https://github.com/kaistAI/Volcano} provided by the authors.
    \item Woodpecker~\cite{yin2023woodpecker}: We used the code\footnote{https://github.com/BradyFU/Woodpecker} provided by the authors. Woodpecker originally uses BLIP2 as the VQA tool, which can limit its performance since BLIP2 is not instruction-tuned. We thus replaced BLIP2 with Instruct-blip and LLaVA-v1.7-7B for a fair comparion with our work. 
\end{itemize}

\paragraph{Evaluation with LLMs:} We used gpt-4-0613 for evaluating \mmhal. GPT-4o was used for evaluation with GAVIE due to lower API costs and claims of similar quality.

\paragraph{Comparative Analysis of Inference Time and API Costs:} We provide the inference time and costs (averaged over 10 examples) for the MME and MMHal-Bench benchmarks using LLaVA-1.5 and the proposed model (Pelican). MME has lower computational costs than MMHal-Bench due to its simpler claims involving single objects or attributes. For Pelican, all tools except the LLM were hosted locally. Pelican's slower performance is due to its stagewise approach; we did not optimize the VQA and detector models for batched requests, which could yield a 2-3x speedup. Additionally, parallelizing tool calls could improve efficiency. The time is also impacted by the number of tool calls, which is smaller for MME (simpler claims) compared to MMHal-Bench. The per-sample API cost varies by dataset, but remains relatively low.

\begin{table}[]
\centering
    \begin{adjustbox}{width=\linewidth}
        \begin{tabular}{cccc}
        \hline
        \textbf{Model} & \textbf{Benchmark} & \textbf{Time} & \textbf{Cost} \\ \hline
        LLaVA-1.5      & MMHal-Bench        & 1.185s        & N/A           \\
        Our Model      & MMHal-Bench        & 5m58s         & \$0.234       \\
        LLaVA-1.5      & MME-existence      & 0.184s        & N/A           \\
        Our Model      & MME-existence      & 2.275s        & \$0.011       \\ \hline
        \end{tabular}
    \end{adjustbox}
    \caption{Inference time and cost comparison of LLaVA-1.5 and our model across MMHal-Bench and MME-existence benchmarks.}
    \label{tab:inference_api}
\end{table}

\paragraph{Additional Evaluation on POPE:} Additionally, we add results on POPE~\cite{li2023evaluating} by selecting 300 examples, following Woodpecker, with 100 samples selected from each of the three settings. We observe consistent improvements over baseline LVLMs, except for a minor accuracy drop in LLaVA-v1.6-7B~\cite{liu2024llavanext}, which was offset by higher precision. We notice the accuracy remains at $91\%$ across all three models with \algo. This consistency is attributed to the Yes/No nature of the questions in the POPE dataset, allowing us to achieve $91\%$ accuracy with LLaVA-v1.6 as the VQA model, regardless of the baseline models' errors. The few failure cases are primarily due to tool failures.

\begin{table}[]
\centering
\begin{adjustbox}{width=\linewidth}
\begin{tabular}{lccc}
\hline
\textbf{Model}         & \textbf{Acc.} & \textbf{Precision} & \textbf{Recall} \\ \hline
llava-v1.6-mistral-7b  & 91.67         & 90.85              & 92.67           \\
+Ours                  & 91.00         & 93.62              & 88.00           \\ 
instructblip-vicuna-7b & 86.00         & 85.06              & 87.33           \\
+Ours                  & 91.00         & 93.62              & 88.00           \\ 
mplug-owl2-llama2-7b   & 80.66         & 75.00              & 92.00           \\
+Ours                  & 91.00         & 93.62              & 88.00           \\ \hline
\end{tabular}
\end{adjustbox}
\caption{Performance comparison of baseline models and ``+Ours" on POPE.}
\label{tab:pope_eval}
\end{table}





\begin{figure*}[t!]
    \centering
        \includegraphics[width=\textwidth]{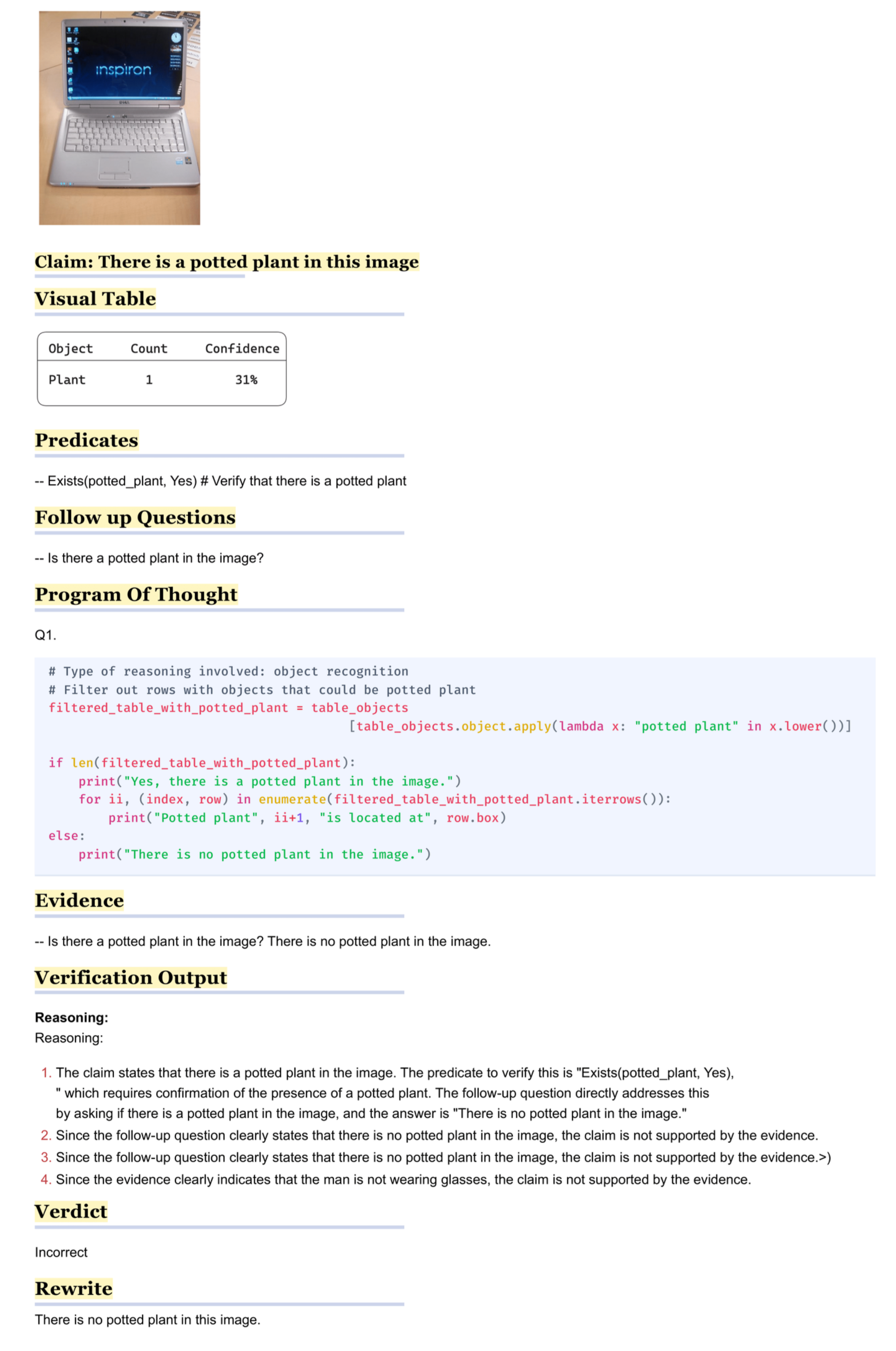}
    \caption{From MME benchmark category existence, the claim is generated from the incorrect answer given by mPlug-OWL. Below shows the entire detection and mitigation pipeline of \algo.}
    \label{fig:example1}
\end{figure*}

\begin{figure*}[t!]
    \centering
        \includegraphics[width=0.87\textwidth]{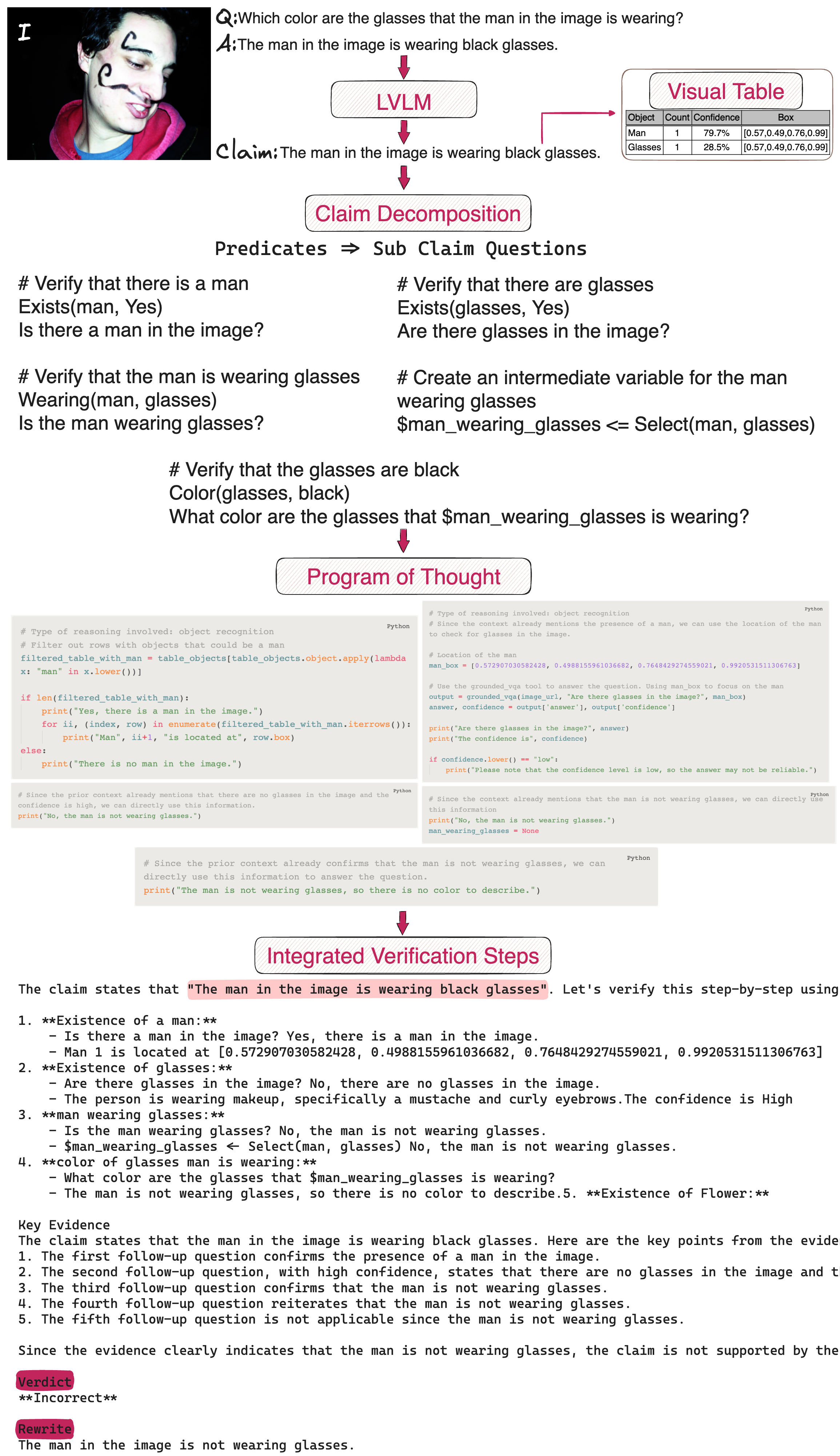}
        \caption{From MMHal-Bench, the claim is generated from the incorrect answer given by LLaVA-v1.5. Below shows the entire detection and mitigation pipeline of \algo.}
    \label{fig:example2}
\end{figure*}


\section{Qualitative Results}
Refer to \autoref{fig:example1} and \autoref{fig:example2} to understand the pipeline depicted for \algo. \autoref{fig:example1} illustrates a simple example from MME, where mPlug-OWL incorrectly detects a potted plant that doesn't exist. We create a predicate to verify the presence of the ``potted plant" followed by its corresponding question and answer. The verification output reasons, leading us to predict the claim as incorrect. We then refine the output by stating, ``There is no potted plant in this image." This example also highlights the brittleness of the tools (refer to limitations in \autoref{sec:limitation}), as ``potted plant" is not a class in YOLO/DETR, so the code defaults to Grounding-DINO (refer to implementation in \autoref{app:implementation}), which detects the plant with low confidence.

\autoref{fig:example2} demonstrates a more complex process, where LLaVA-v1.5 misidentifies a face painting as glasses. The detection tools find a ``man" with high confidence via YOLO, and ``glasses" are detected again via Grounding-DINO with low confidence. Next, we prepare a list of predicates and follow-up questions. Next, we show the Python code for each follow-up question. Finally, the evidence gathered is used for reasoning to predict the claim as incorrect.

\section{Licensing Information}
The images used in our paper are sourced from MSCOCO, Visual Genome and OpenImages that are under different Creative Commons licenses. Images \href{https://farm4.staticflickr.com/2503/3828840628_e625c9beb0_o.jpg}{\autoref{fig:model}}, \href{https://c3.staticflickr.com/4/3169/2653797190_f42908d6e5_o.jpg}{left in \autoref{fig:demo}}, \href{https://farm4.staticflickr.com/41/83372724_a17cceb078_o.jpg}{\autoref{fig:example2}} are licensed under CC BY 2.0, which allows for sharing and adaptation with appropriate attribution. \href{https://farm3.staticflickr.com/2325/5723995240_7382f64328_z.jpg}{center in \autoref{fig:demo}}, \href{http://farm4.staticflickr.com/3613/3743850787_80d30fd5df_z.jpg}{\autoref{fig:example1}}  are under CC BY-NC 2.0, permitting non-commercial use with attribution. Image on the right in \autoref{fig:demo} is taken from \href{https://homes.cs.washington.edu/~ranjay/visualgenome/about.html}{Visual Genome} dataset which is under CC BY 4.0 which allows for sharing and adaptation with appropriate attribution. Users should refer to the specific license links provided for each image to ensure compliance with usage terms.

\begin{figure*}[t!]
    \centering
        \includegraphics[width=\textwidth]{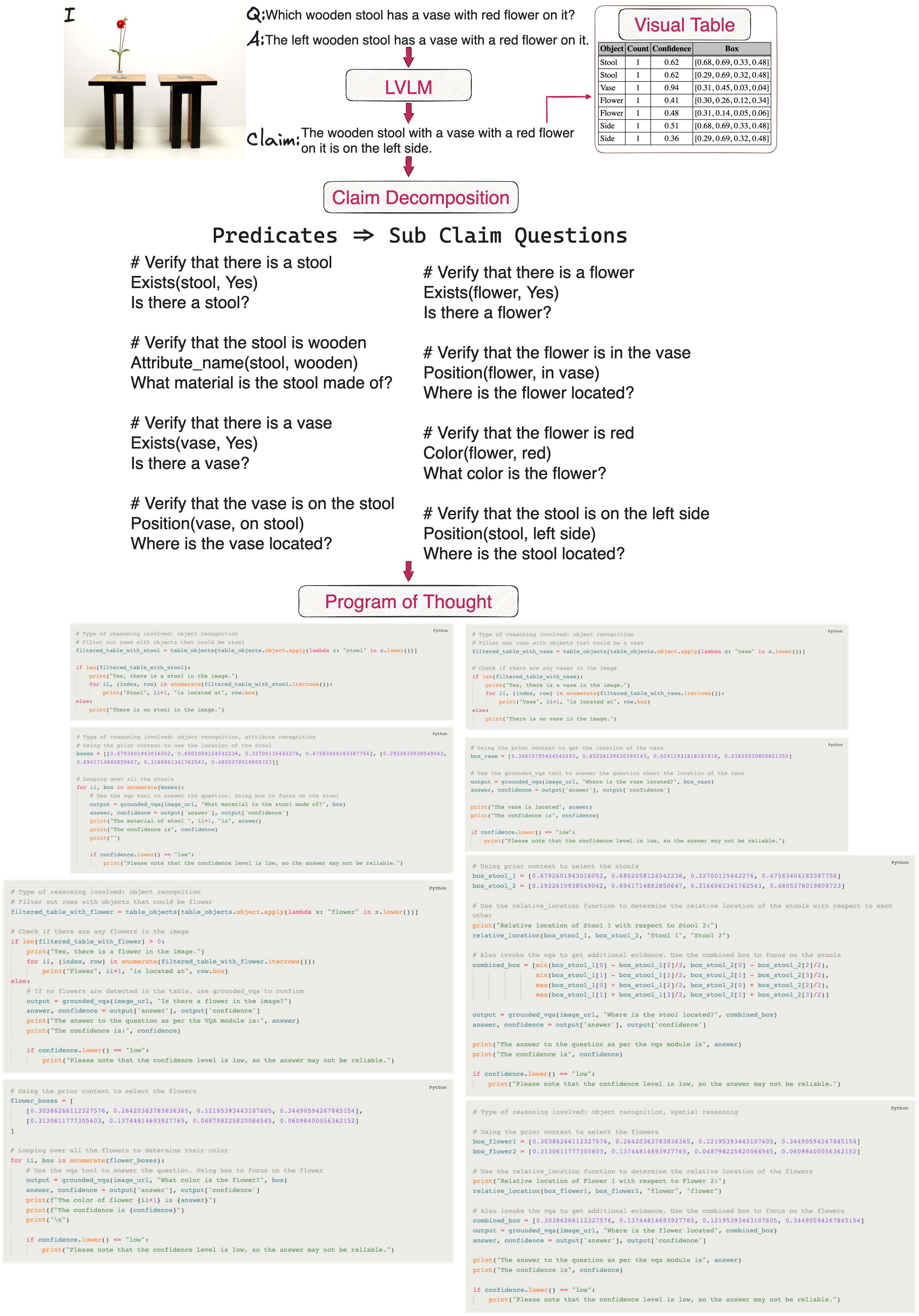}
    \phantomcaption
\end{figure*}
\begin{figure*}[t!]
\ContinuedFloat
    \centering
        \includegraphics[width=\textwidth]{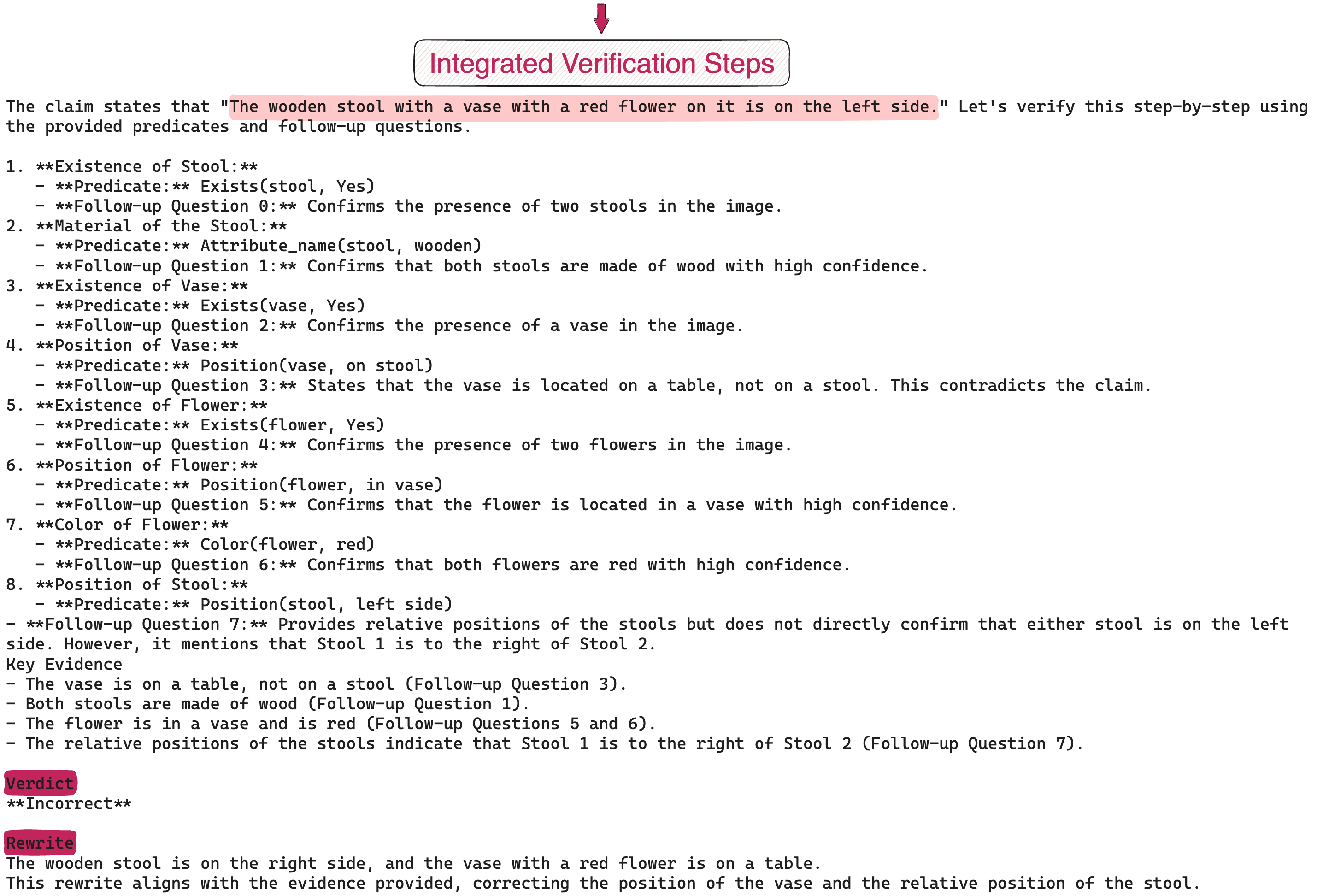}
\caption{In MMHal-Bench, the claim is generated from the correct answer provided by LLaVA-v1.5. Below is the complete detection and mitigation pipeline of \algo. This is a negative example where \algo makes an error. Although the evidence collected was largely accurate, there was an inconsistency between identifying a table and a stool, which likely arose from the brittle nature of the object detection tool. In the evidence analysis section, the relative position was inaccurately assessed, leading to an incorrect final verdict.}
    \label{fig:example4}
\end{figure*}

\end{document}